\documentclass[conference]{IEEEtran}

\pdfoutput=1

\usepackage[utf8]{inputenc}
\usepackage[T1]{fontenc}

\usepackage[final]{microtype}            
\usepackage{enumitem}                    
\usepackage{balance}                     

\usepackage{cite}
\usepackage{amsmath,amssymb,amsfonts}
\usepackage{algorithmic}
\usepackage{graphicx}
\usepackage[table]{xcolor}      
\usepackage{textcomp}

\usepackage[hidelinks]{hyperref}

\renewcommand{\arraystretch}{0.95}
\setlist[itemize]{noitemsep,topsep=2pt,leftmargin=*,parsep=0pt,partopsep=0pt}
\setlist[enumerate]{noitemsep,topsep=2pt,leftmargin=*,parsep=0pt,partopsep=0pt}

\IEEEoverridecommandlockouts
\IEEEaftertitletext{\vspace{-0.6\baselineskip}}

\renewenvironment{thebibliography}[1]{%
  \begin{oldthebibliography}{#1}%
  \setlength{\itemsep}{0pt}%
  \setlength{\parskip}{0pt}%
  \setlength{\parsep}{0pt}%
}{\end{oldthebibliography}}

\def\BibTeX{{\rm B\kern-.05em{\sc i\kern-.025em b}\kern-.08em
    T\kern-.1667em\lower.7ex\hbox{E}\kern-.125emX}}

\begin{document}

\title{Zero-Shot Referring Expression Comprehension via Vision-Language True/False Verification}

\author{%
  \IEEEauthorblockN{Jeffrey Liu \qquad Rongbin Hu}
  \IEEEauthorblockA{\textit{mycube.tv}, San Francisco, U.S.A.}
}

\maketitle

\begin{abstract}
Referring Expression Comprehension (REC) is usually addressed with task-trained grounding models. We show that a zero-shot workflow, without any REC-specific training, can achieve competitive or superior performance. Our approach reformulates REC as box-wise visual–language verification: given proposals from a COCO-clean generic detector (YOLO-World), a general-purpose VLM independently answers \texttt{True/False} queries for each region. This simple procedure reduces cross-box interference, supports abstention and multiple matches, and requires no fine-tuning. On RefCOCO, RefCOCO+, and RefCOCOg, our method not only surpasses a zero-shot GroundingDINO baseline but also exceeds reported results for GroundingDINO trained on REC and GroundingDINO+CRG. Controlled studies with identical proposals confirm that verification significantly outperforms selection-based prompting, and results hold with open VLMs. Overall, we show that workflow design, rather than task-specific pretraining, drives strong zero-shot REC performance.
\end{abstract}

\begin{IEEEkeywords}
Referring expression comprehension, large language modeling inference, zero-shot reasoning
\end{IEEEkeywords}

\section{Introduction}

\begin{figure*}[!t]
    \centering
    \includegraphics[width=\textwidth]{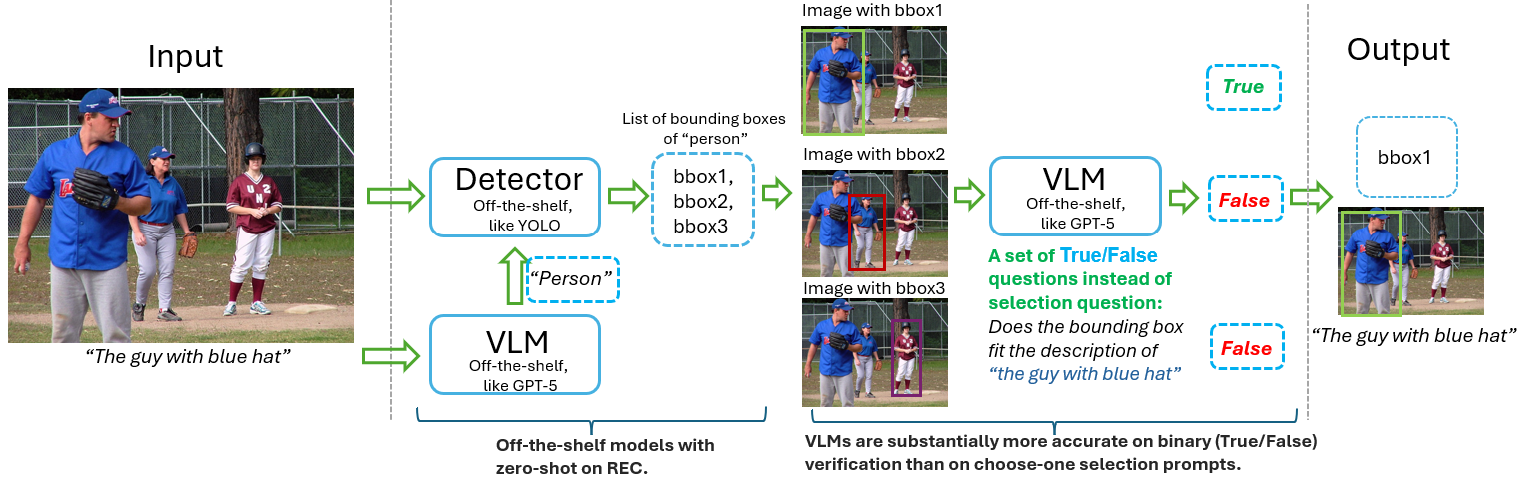}
    \caption{The concept behind our verification-first workflow, which enables outstanding accuracy in a training-free zero-shot setting.}
    \label{fig:workflow}
\end{figure*}

Referring Expression Comprehension (REC) \cite{qiao2020rec_survey} is the task of locating the specific object described by a natural language expression (e.g., `the small red mug on the left') in a given image. REC supports interactive perception and grounding for applications such as camera auto-directing and framing in live streaming, video content editing, and voice-directed robotic/device control. The standard output is a single bounding box that identifies the referent. 

Despite strong progress in large vision–language models (VLMs), e.g., CLIP\cite{radford2021clip}, LLaVA\cite{liu2023llava}, GPT-5\cite{openai_gpt5_system_card_2025}, REC remains difficult. The task demands fine-grained language understanding and grounded visual reasoning to select one instance among many look-alike objects. Descriptions blend attributes, parts, and spatial/ordinal cues, often in scenes with occlusion and scale changes, so that close similarities are common. Furthermore, off-the-shelf VLMs without REC training rarely produce accurate bounding boxes as they are not calibrated for instance-level localization. 

In the supervised setup, models learn the REC mapping directly from annotated referring datasets. Approaches include end-to-end grounding architectures and cross-modal matching heads that predict a box from an image–text pair. Text-conditioned detectors such as GroundingDINO\cite{liu2023groundingdino} show substantial gains when trained with RefCOCO-style supervision. Supervised large multimodal models, including CogVLM\cite{wang2024cogvlm} variants finetuned for grounding, currently define the-state-of-the-art under this regime.

Another common setup uses grounding-trained proposal generators with inference-time decision logic. A text-conditioned detector (often GroundingDINO) produces language-aware region proposals with bounding boxes, and a separate module refines or selects among them at test time. CRG\cite{wan2024crg} is a representative example that re-scores proposals with a VLM. Although the final stage operates without additional training, the overall pipeline is not zero-shot at the system level, because the proposal source has been trained on REC-style data.

Finally, in \textbf{zero-shot workflows}\cite{frome2013devise}\cite{xian2019zsl}, both components are off-the-shelf: a generic detector supplies candidate boxes, and a VLM provides the decision logic via prompting. Neither of the models is trained on REC. The decision logic is commonly based on global ranking of the top-K candidates. Performance is bounded by proposal recall of the generic detector. The workflow, including prompt design, decomposition into attributes and relations, and tie-breaking, becomes the primary lever for accuracy and robustness. Our method targets this zero-shot workflow regime.

Our method treats REC as box-wise verification on the original image. A generic detector proposes candidates. For each proposal, we render the image with only that box drawn and ask a VLM a single binary question: \emph{does this box match the description?} The VLM returns \texttt{True/False} without scores. This first pass shrinks the candidate set. If exactly one box is \texttt{True}, we return it. If multiple boxes are \texttt{True}, we render the image with only those \texttt{True} boxes overlaid together and ask the VLM to select the one that best fits the description, with all \texttt{False} boxes being excluded from this step. If all boxes are \texttt{False}, we fall back to a global selection prompt over the full proposal set.

On RefCOCO\cite{yu2016refcoco}, RefCOCO+, and RefCOCOg\cite{mao2016refcocog}, our verification-first workflow with GPT-5 achieves ACC@0.5 scores of 79.3\%, 74.2\%, and 72.4\%, respectively. This marks a substantial improvement over the zero-shot GroundingDINO baseline without RefCOCO training (50.4\%, 51.4\%, 60.4\%). Our approach sets a new state of the art for these benchmarks under zero-shot conditions. Here, ‘zero-shot’ means that no model is trained on the REC task, the detector is COCO-clean (trained without COCO\cite{lin2014coco} data), and the VLM is used strictly off-the-shelf—though its pretraining corpus may include COCO images, which is not explicitly known. It also surpasses the reported performance of GroundingDINO with RefCOCO training by 7.2\% on average. Compared with a single-shot selection approach that uses hand-crafted prompts over the same proposals and VLMs, our proposed \texttt{True/False} verification workflow improves ACC@0.5(\%) by 12.3\% on average across the three benchmarks and two VLMs. 

Zero-shot REC is valuable for both practicality and scientific insight. Practically, it allows developers to build solutions from widely available, off-the-shelf components instead of relying on task-specific models that require curated labels, specialized training, and repeated re-tuning as domains shift. More importantly, zero-shot REC offers a clean testbed for workflow design: can a carefully constructed procedure with generic models to solve a specialized task? Conceptually, REC decomposes into two basic capabilities: (1) visual–language comprehension and reasoning, and (2) deriving a bounding box for a given region. If these fundamentals can be combined to solve REC, it suggests that other composite tasks may likewise be addressed modularly by reusing general-purpose models for their constituent subtasks. 

Our results echo the trends in LLMs where prompt design, tool use, and staged reasoning often replace fine-tuning. By showing that \texttt{True/False} verification narrows and often resolves the decision problem, we present a template that extends beyond REC to other complex vision tasks (or even generic LLM based tasks). Our central insight includes:
\begin{itemize}
  \item \textbf{Verification over comparison.}
  Modern vision--language models reason more reliably when prompted with a single, concrete hypothesis rather than a comparative choice among many alternatives. Each query of \emph{box-wise verification} isolates a single candidate and removes cross-candidate coupling, order effects, and prompt entanglement, thereby minimizing interference.
  \item \textbf{Selection as atomic checks.}
  Any multi-candidate selection can be expressed as a set of independent \texttt{True/False} tests. Running verification once per proposal converts a joint decision into parallel atomic decisions and yields a pruned candidate set. Even in the multi-\texttt{True} case, this reduction sharply shrinks the search space and empirically improves the final selection’s accuracy and stability.

 \end{itemize}

\section{Method}
\label{sec:method}

\begin{table}[t]
\caption{Zero-shot, verification-first REC}
\label{tab:rec-fallback}
\centering
\begin{minipage}{0.96\columnwidth}
\footnotesize
\begin{tabular}{@{}p{0.03\columnwidth}p{0.90\columnwidth}@{}}
1. & Input image $I$, description $s$, detector $D$, VLM $F$. \\
2. & $c \leftarrow F$ infer class from $s$. \\
3. & $\mathcal{B} \leftarrow D(I,c)$ (candidate boxes). \\
4. & For each $b_i \in \mathcal{B}$: overlay $\tilde{I}_i$ and query $F(\tilde{I}_i,s)\!\to\! y_i\in\{\texttt{True},\texttt{False}\}$. \\
5. & $\mathcal{T}\leftarrow \{i \mid y_i=\texttt{True}\}$. If $|\mathcal{T}|=1$, return that box. \\
6. & If $|\mathcal{T}|>1$, overlay \& index only $\{b_i\}_{i\in\mathcal{T}}$; ask $F$ to pick best; return it. \\
7. & If $\mathcal{T}=\varnothing$, overlay \& index all boxes; ask $F$ to pick best; if “none,” abstain. \\
\end{tabular}
\end{minipage}
\label{tab:algorithm}
\end{table}

\subsection{Problem Setup and Zero-Shot Setting}
REC takes an image and a description and returns the best-matching bounding box (or abstains if none fits). We use a zero-shot workflow with only off-the-shelf models: a \emph{basic, class-conditioned detector without grounding capability} to propose boxes, and \emph{a general-purpose VLM} to verify them.\footnote{Zero-shot means no REC training; the detector is \emph{COCO-clean}; the VLM is used as-is and may or may not have seen COCO during pretraining.} This shifts task specialization from model weights to the procedure.

\subsection{Verification-First Workflow}
Our workflow reformulates REC as \emph{box-wise visual--language verification} and resolves the final choice with a lightweight VLM-driven selector.  As shown in Figure~\ref{fig:workflow} and Table~\ref{tab:algorithm}, the workflow has the following steps: 

\begingroup
\renewcommand{\theenumi}{\alph{enumi}}
\renewcommand{\labelenumi}{(\theenumi)}
\begin{enumerate}
  \item \textbf{Class identification.}
  From the natural-language description, ask the VLM to name the most relevant object class (e.g., person, dog, car). Use this class to focus the detector.

  \item \textbf{Class-conditioned proposals.}
  Run the detector on the image for that class to produce a set of candidate bounding boxes.

  \item \textbf{Box-wise verification.}
  For each candidate box, draw it on the image and ask the VLM: ``Does the description apply to this box?'' Record a \texttt{True}/\texttt{False} judgment for each box.

  \item \textbf{Decision rule (with abstention).}
  \begin{itemize}
    \item If exactly one box is labeled \texttt{True}, output that box.
    \item If multiple boxes are labeled \texttt{True}, show only those boxes (with indices) and ask the VLM to choose the best-fitting one; output it.
    \item If all boxes are labeled \texttt{False}, show all boxes (with indices) and ask the VLM to choose the best match; if it replies ``none,'' abstain.
  \end{itemize}
\end{enumerate}
\endgroup

\begin{figure*}[!t] 
  \centering

  \begin{minipage}{0.32\textwidth}
    \centering
    \includegraphics[width=\linewidth]{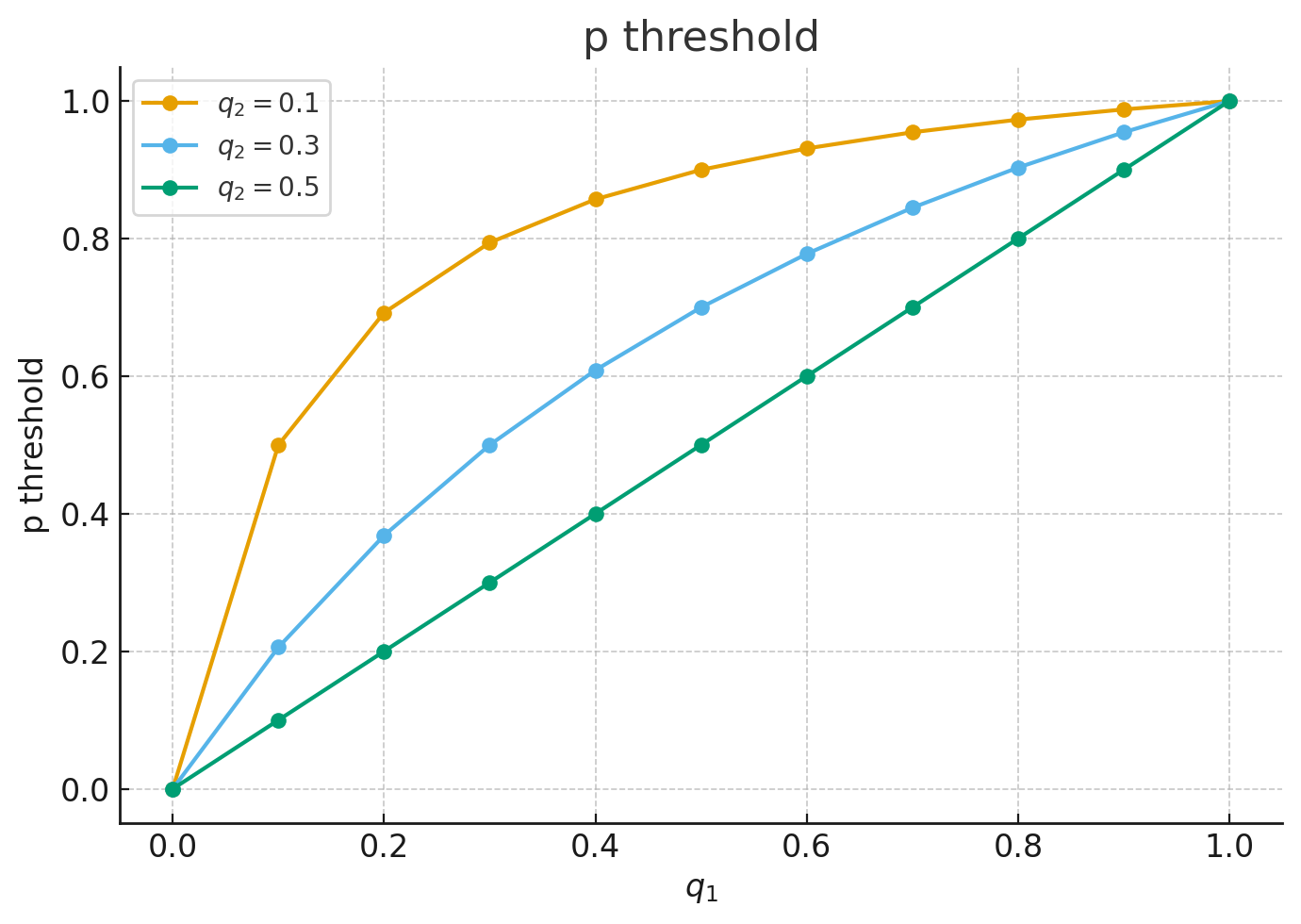}
    \vspace{0.25ex}
    \small (a) $p$ threshold vs.\ $q_1$ and $q_2$.
  \end{minipage}\hfill
  \begin{minipage}{0.32\textwidth}
    \centering
    \includegraphics[width=\linewidth]{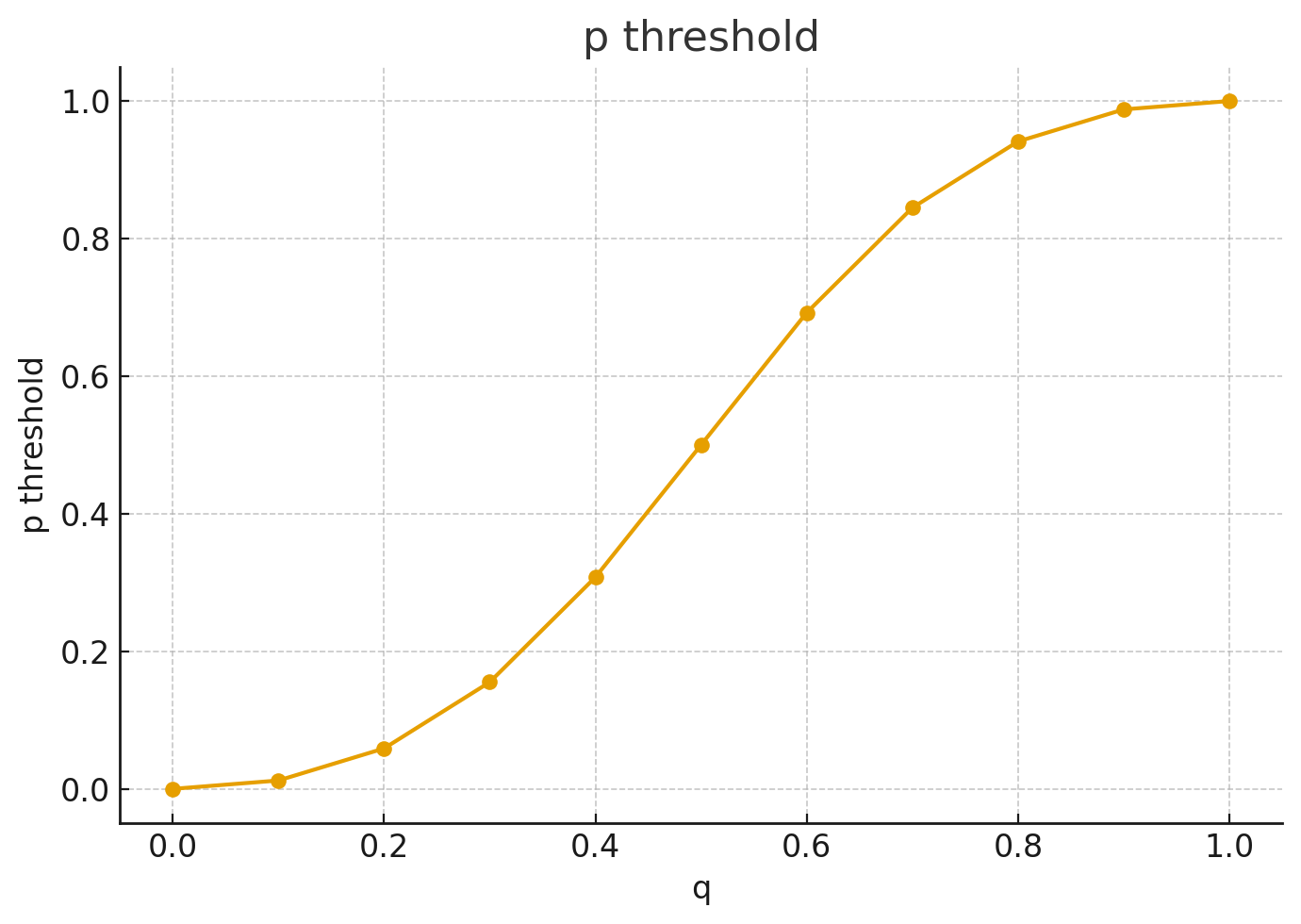}
    \vspace{0.25ex}
    \small (b) $p$ threshold as a function of $q$.
  \end{minipage}\hfill
  \begin{minipage}{0.32\textwidth}
    \centering
    \includegraphics[width=\linewidth]{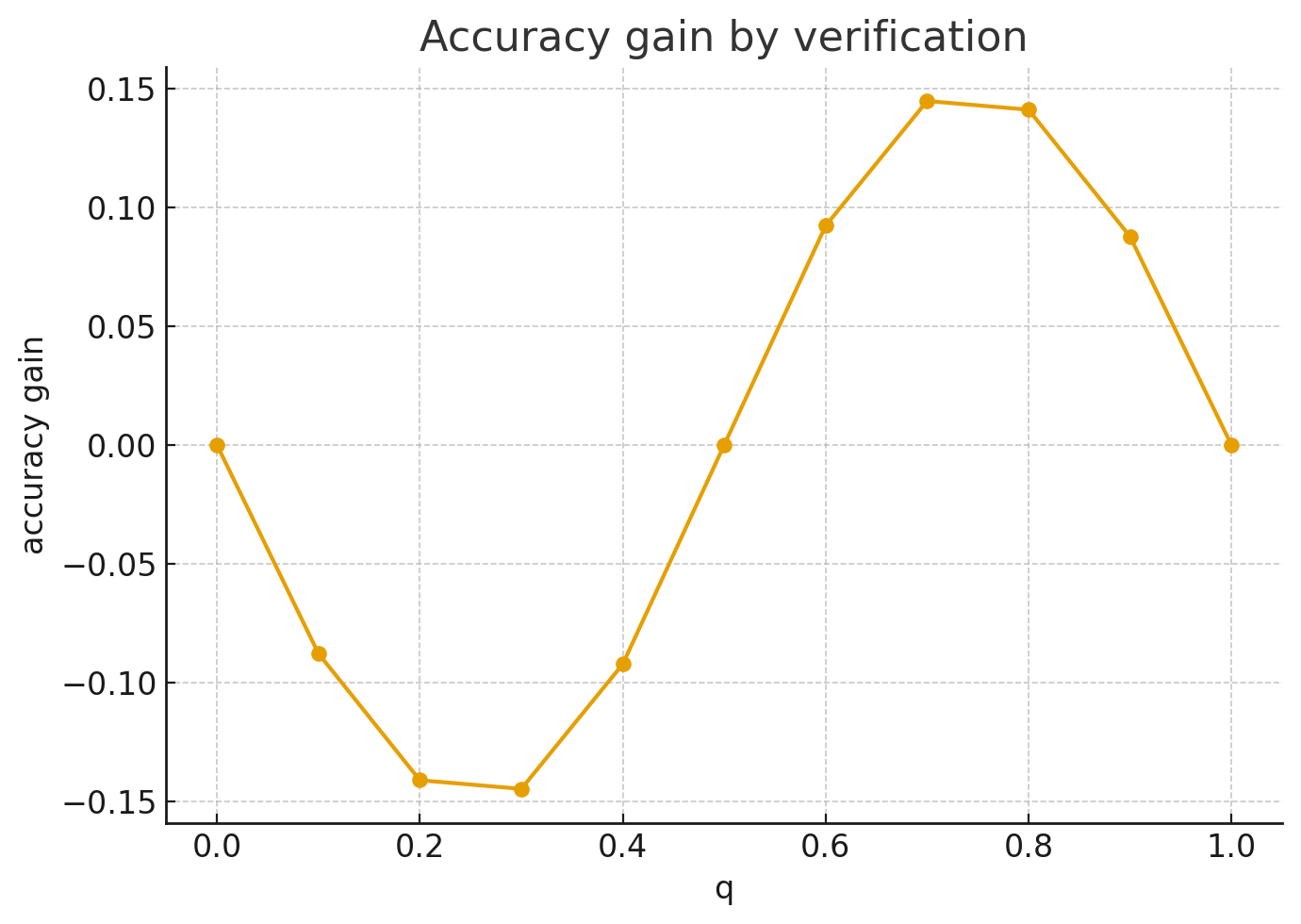}
    \vspace{0.25ex}
    \small (c) Accuracy gain by verification vs selection.
  \end{minipage}

  \caption{Threshold and gain analyses derived from the two-candidate model (Sec.~\ref{sec:analysis}).}
  \label{fig:threshold_gain_row}

  \vspace{12pt}

  \begin{minipage}{\textwidth}
    \centering
    \footnotesize
    \setlength{\tabcolsep}{5.5pt}
    \renewcommand{\arraystretch}{1.0}

    \refstepcounter{table}
    \textbf{TABLE \thetable}\\[2pt]
    REC benchmarking (ACC@0.5 (\%)).\label{tab:bench}
    \vspace{4pt}

    \begin{tabular}{l l ccc ccc cc}
      \hline
      Regime & Method &
      \multicolumn{3}{c}{RefCOCO} &
      \multicolumn{3}{c}{RefCOCO+} &
      \multicolumn{2}{c}{RefCOCOg} \\
      \cline{3-10}
      & & Val & TestA & TestB & Val & TestA & TestB & Val & Test \\
      \hline
      Supervised                    & CogVLM (REC-trained)                    & 92.6 & 94.3 & 91.5 & 85.2 & 89.6 & 79.8 & 88.7 & 89.4 \\
      Supervised                    & GroundingDINO (REC-trained)             & — & 77.3 & 72.5 & — & 72.0 & 59.3 & — & 66.3 \\
      \hline
      Workflow with REC             & GroundingDINO + CRG (REC-trained rerank)& — & 81.6 & 73.2 & — & 77.0 & 60.0 & — & 69.6 \\
      \hline
      Strict zero-shot              & GroundingDINO (zero-shot baseline)      & 50.4 & 57.2 & 43.2 & 51.4 & 57.6 & 45.8 & 60.4 & 59.5 \\
      \hline
      Zero-shot VLM (baseline)      & GPT-5 (organic / vanilla selection)    & 11.7 & — & — & 10.1 & — & — & 12.5 & — \\
      \hline
      Zero-shot on REC (ours)       & Selection (LLaVA, single-shot)          & 34.7 & — & — & 34.6 & — & — & 44.4 & — \\
      Zero-shot on REC (ours)       & Verification-first (LLaVA)               & 44.6 & — & — & 42.3 & — & — & 50.7 & — \\
      Zero-shot on REC (ours)       & Selection (GPT-5, single-shot)          & 70.1 & 73.5 & 65.4 & 66.7 & 68.3 & 60.2 & 69.7 & 68.4 \\
      Zero-shot on REC (ours)       & Selection (GPT-5, majority voting)          & 71.7 & — & — & 67.4 & — & — & 69.9 & — \\
      \textbf{Zero-shot on REC (ours)} & \textbf{Binary verification (GPT-5)}           & \textbf{79.3} & \textbf{85.6} & \textbf{70.4} & \textbf{74.2} & \textbf{80.4} & \textbf{65.2} & \textbf{72.4} & \textbf{71.5} \\
      \hline
    \end{tabular}
  \end{minipage}
\end{figure*}

Rather than asking the VLM to choose one box among many, we verify each candidate independently with a binary \texttt{True/False} query. Although the binary verification workflow may appear simple and superficially similar to selection-based prompting, it behaves quite differently in practice. We expect the performance improvements to stem from three mechanisms:
\begin{itemize}
    \item \textbf{Reduced cross-box interference:} Binary verification workflow decouples candidates and reduces cross-box interference, especially when the detector generates many proposals. Focusing the model on a single highlighted region concentrates its reasoning and limits distraction from nearby objects.
    \item \textbf{Built-in error control:} Binary verification also provides a simple yet strong error-control signal: when exactly one box is labeled \texttt{True}, without being told how many positives to expect, the VLM has singled out a unique region that satisfies the description. This outcome is far more likely to reflect genuine comprehension and reasoning than chance.
    \item \textbf{Pruned candidates:} When multiple boxes are \texttt{True}, the set of candidates is already pruned, allowing a lightweight second-stage tie-break over a smaller and cleaner pool.
\end{itemize} 

\section{Analysis}
\label{sec:analysis}

Consider two proposals: bounding box~1 is correct and bounding box~2 is a distractor.
Under a selection prompt, let us define $p$ as the probability of selecting bounding box 1, so
\begin{equation}
A_{\text{sel}} = p.
\end{equation}
Under box-wise \texttt{True/False} verification, let us define $q_1$ and $q_2$ as probabilities of bounding box 1 and bounding box 2 being labeled \texttt{True}, respectively. 
When both are \texttt{True} or both are \texttt{False}, it falls back to the selection setup. Then the accuracy of the verification approach is
\begin{equation}
\label{eq:aver}
A_{\text{ver}}
= q_1(1-q_2) + q_1 q_2 p + (1-q_1)(1-q_2)p.
\end{equation}

To determine when selection outperforms verification ($A_{\text{sel}}\ge A_{\text{ver}}$), we can have:
\begin{equation}
\label{eq:threshold}
p \;\ge\; 1 \;-\; \frac{q_2(1-q_1)}{\,q_1(1-q_2)+q_2(1-q_1)\,}.
\end{equation}
Thus selection must exceed the threshold in \eqref{eq:threshold} to beat verification. 

As shown in Figire~\ref{fig:threshold_gain_row} (a), for any fixed $q_2<0.5$, the threshold of $p$ is an increasing and concave function of $q_1$ and lies above the identity line $p=q_1$ for $0<q_1<1$. Hence, to match the accuracy of verification, the selection method must achieve $p>q_1$ (i.e., be more reliable than the verifier’s true-positive rate). Design-wise, we aim to keep $q_2<0.5$. Note that $q_2=0.5$ corresponds to random guessing on the distractor, so effective prompting/overlays should push $q_2$ well below this. With a reasonably capable VLM and small $q_2$, the verification succeeds more easily, and selection must be correspondingly stronger to keep up.

To further simplify, assume $q_1=q$ and $q_2=1-q$. As shown in Figure~\ref{fig:threshold_gain_row} (b), $p$ threshold lies above the identity line $p=q$ for $q>0.5$. To match verification, the selection method must achieve a probability $p$ strictly higher than the verifier’s true-positive rate $q$. Figure~\ref{fig:threshold_gain_row} (c) plots the gap between $p$ threshold and $q$, which peaks at about $0.145$ near $q\!\approx\!0.7$. In other words, when the verifier labels the correct box \texttt{True} with probability $q\!\approx\!0.7$ (and the distractor with $1-q\!\approx\!0.3$), selection would need $p\!\approx\!0.845$—roughly a $+0.15$ absolute increase—to achieve the same accuracy.

One might view this comparison as unfair: the selection baseline uses a single run, whereas our verification-first pipeline uses at least two rounds (and a third only to break ties). In principle, we could also run the selection procedure three times and take a majority vote; under an i.i.d.\ assumption this yields an effective accuracy of \(3q^2 - 2q^3\) and would require us to compare against the condition \(p > 3q^2 - 2q^3\), which is stricter than the threshold in Eq.~\eqref{eq:threshold}. However, two caveats matter. First, majority-vote selection requires three rounds, while for large \(q\) the verification-first scheme typically concludes in two; thus the compute budgets are not apples-to-apples. Second, repeated calls to the same VLM on the same instance are rarely i.i.d.\ --- especially at the low temperatures commonly used for selection --- so multiple runs often reproduce the same output, making majority voting little better than a single run. (This assumption is validated in Table~\ref{tab:bench}). By contrast, verification-first queries different regions/objects across rounds, making the i.i.d.\ assumption more plausible. For these reasons, we evaluate the selection baseline with a single run in this analysis.

The two-box analysis omits two sources of advantage in practice. First, \emph{cross-box interference} is suppressed under verification because each query presents a single highlighted region. This typically \emph{raises} the verifier’s true-positive rate while \emph{not} improving the selector, widening the gap required for selection to match verification.
Second, with more than two proposals, verification induces \emph{pruning}, while the two-box model cannot reflect the gain as there is no meaningful shrinkage when $n=2$.

\section{Experiments}
\label{sec:experiments}
We evaluate on RefCOCO, RefCOCO+, and RefCOCOg, i.e., the standard REC benchmarks derived from MS-COCO, each providing image-expression pairs. Following RefCOCO conventions, a detection is correct if its Intersection-over-Union (IoU) with the ground-truth box exceeds 0.5. We therefore report \textbf{ACC@0.5}. We consider the following approaches in our experiments for comparison: 
\begin{itemize}
  \item \textbf{CogVLM (supervised).} A vision–language model trained with REC supervision and evaluated in the standard supervised regime.
  \item \textbf{GroundingDINO (REC-trained).} GroundingDINO fine-tuned on REC datasets to directly localize the referred object.
  \item \textbf{GroundingDINO + CRG (REC-trained workflow).} A re-ranking workflow that scores proposals from REC-trained GroundingDINO using language/context cues to select the final box.
  \item \textbf{GroundingDINO (zero-shot baseline).} A pretrained text-conditioned detector applied without any REC fine-tuning to output the top-1 box.
  \item \textbf{GPT-5 (vanilla REC).} Off-the-shelf GPT-5 is shown the image with description and asked to generate the best-matching bounding box in a single prompt (no REC training).
  \item \textbf{Verification-first (LLaVA, ours).} Use YOLO-world\cite{cheng2024yoloworld}, a COCO-clean basic class-conditioned detector (no grounding), proposes boxes, and off-the-shelf and open LLaVA verifies each box with a \texttt{True/False} query, with abstention/tie-breaks as needed (no REC training).  
  \item \textbf{Verification-first (GPT-5, ours).}   Same verification pipeline as above but using GPT-5 instead of LLaVA.
  \item \textbf{Selection (LLaVA, ours).} Control variant that uses the same proposals and an open VLM (LLaVA) issues a single-shot selection prompt rather than box-wise verification.
  \item \textbf{Selection (GPT-5, single-shot, ours).} Same selection pipeline as above but using GPT-5 instead of LLaVA.  
  \item \textbf{Selection (GPT-5, majority voting, ours).} Same selection pipeline as above, using GPT-5, but using majority voting of three runs. Temperature is configured to 1.0.  
\end{itemize}

We use YOLO-World as a basic class-conditioned detector (no grounding), trained \emph{without} COCO for dataset cleanliness. This COCO-clean choice comes with a modest cost, i.e., approximately 10 percentage points mAP@0.5 lower than comparable COCO-trained YOLO variants on detection tasks. For VLMs, we evaluate GPT\text{-}5 and LLaVA-vicuna-13b off-the-shelf. A “organic” GPT\text{-}5 baseline attains less than 15\% ACC@0.5 on these benchmarks, suggesting that it is \emph{not} REC-trained. The strong numbers we obtain stem from the workflow rather than hidden task-specific pretraining.  

As shown in Table~\ref{tab:bench}, except for the fully supervised CogVLM, our \emph{verification-first (GPT\text{-}5)} approach outperforms most other approaches across RefCOCO/RefCOCO+/RefCOCOg. Using the \emph{same} detector, VLM, and proposal sets, verification consistently beats selection, including the selection with majority voting, by about 5 to 10 absolute points across most of the splits, underscoring that workflow—not just model choice—drives the gains. We also observe that GPT\text{-}5 surpasses LLaVA in both verification and selection, indicating that VLM capability is critical in zero-shot REC. In summary: (i) a carefully designed workflow, like the \texttt{True/False} verification workflow, can deliver large improvements even with identical components, and (ii) fundamental models (a basic, non-grounding detector and a general-purpose VLM) can be composed to solve a composite task like REC without task-specific training—though the strength of each component still matters.

\section{Conclusion}
We present a verification-first, zero-shot REC workflow that achieves high performance. Extensive experiments show consistent gains over single-shot selection and over REC-trained GroundingDINO(+CRG). The results indicate workflow design, rather than task-specific pretraining, drives the improvement. Future work will extend to other vision tasks.

\balance

\end{document}